# A Nearable Soft Mat Based on Distributed Optical Fiber Sensing for Physiological Monitoring

V. Lavorgna, Graduate Student Member, IEEE, M. Pulcinelli, Graduate Student Member, IEEE, A. Polimadei, R. D'Amato, C. Massaroni, Member, IEEE, M.A. Caponero, E. Schena , Senior Member, IEEE, D. Lo Presti, Member, IEEE

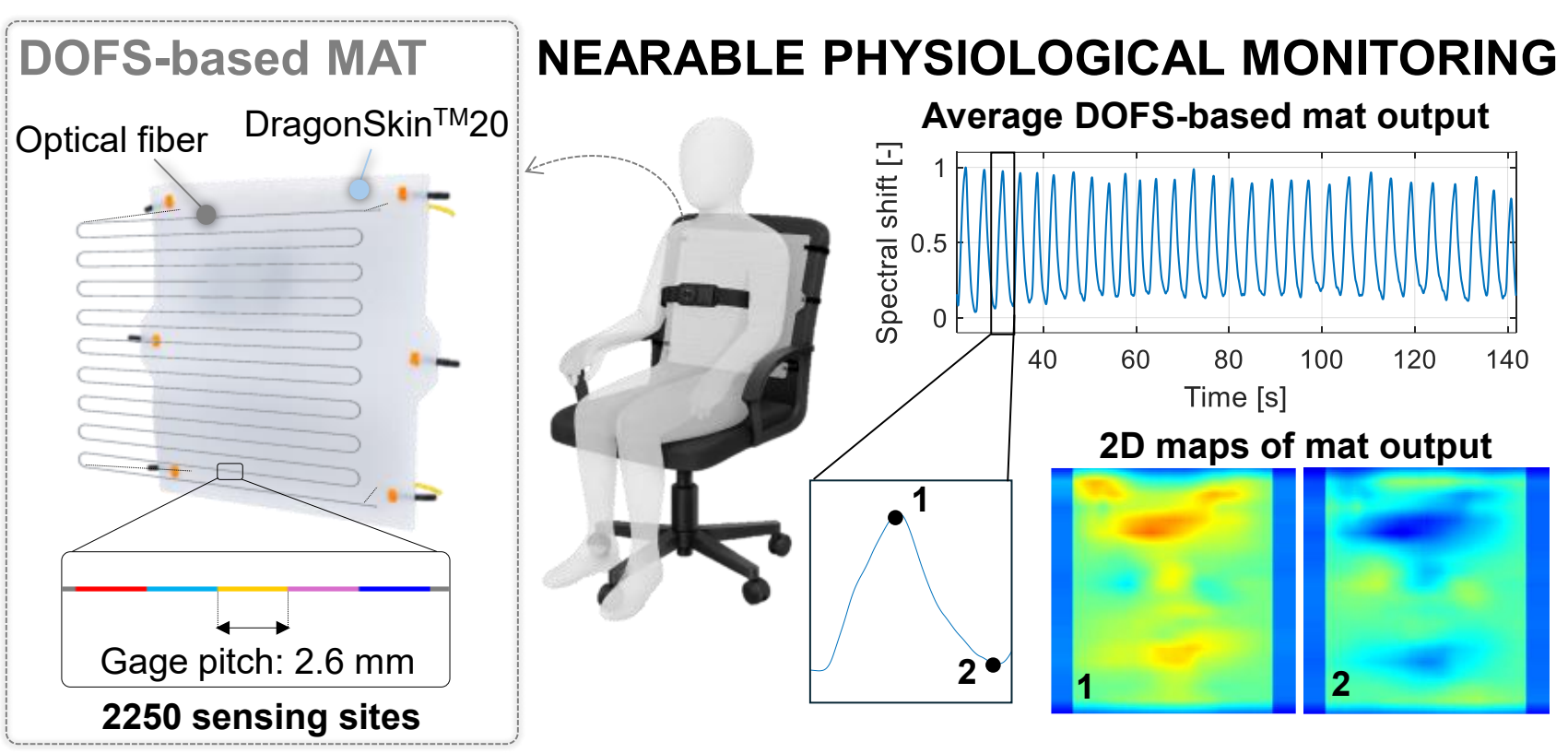

**Abstract -** Distributed optical fiber sensing (DOFS) combines the advantages of fiber-optic sensors, including flexibility, small size, immunity to electromagnetic interference, and high metrological performance, with the capability to transform a single optical fiber into a continuous sensing element for spatially resolved mechanical measurements. Optical frequency domain reflectometry (OFDR), based on Rayleigh backscattering, enables high-spatial-resolution DOFS measurements, broadening the range of potential sensing applications. However, OFDR-based DOFS remains largely unexplored for biomedical applications, despite the need for sensitive, spatially resolved, and conformable sensing interfaces. This study presents a soft DOFS-based mat as a large-area interface for physiological monitoring. A single-mode optical fiber was embedded in a flexible silicone matrix and arranged in a serpentine layout to distribute sensing over the mat surface. With a gage pitch of 2.6 mm, the system provided 2250 sensing sites across the active area at a sampling frequency of 50 Hz. The mat was assessed on six healthy volunteers in a seated nearable configuration on the backrest of a standard office chair. The distributed output enabled two-dimensional mapping of the mat response, reflecting back–mat mechanical coupling and cardiorespiratory-induced perturbations. Respiratory rate and heart rate were therefore estimated and compared with a reference wearable system. The maps revealed physiologically coherent spatial and temporal patterns, while the estimated rates showed good agreement with the reference measurements. These results demonstrate the feasibility of combining large-area distributed sensing, spatial mapping, and quantitative cardiorespiratory monitoring within a DOFS-based soft nearable interface.

**Index Terms**—distributed optical fiber sensing, optical frequency domain reflectometry, fiber optic sensors, soft mat, cardiorespiratory monitoring, silicone matrix

V. Lavorgna, M. Pulcinelli, C. Massaroni, D. Lo Presti, and E. Schena are with the Unit of Measurements and Biomedical Instrumentation, Department of Engineering, Università Campus Bio-Medico di Roma, 00128 Rome, Italy.
A. Polimadei, R. D'Amato, and M.A. Caponero are with the Photonics Micro- and Nanostructures Laboratory, ENEA Research Center of Frascati, 00044 Frascati, Italy.
C. Massaroni, D. Lo Presti, and E. Schena are part of Fondazione Policlinico Universitario Campus Bio-Medico, 00128 Rome, Italy.

## I. INTRODUCTION

In recent years, fiber optic sensors (FOSs) have become enabling technologies for the development of smart sensing systems, owing to their compactness, flexibility, immunity to electromagnetic interference, and excellent metrological properties [1]. These features allow FOSs to measure several physical quantities, including strain, temperature, force, pressure, and vibrations, with high sensitivity and reliability [2], [3]. Among FOS technologies, distributed optical fiber sensing (DOFS) represents a particularly attractive approach, as it turns a single optical fiber into a continuous sensing medium capable of providing spatially distributed measurements along its length [4]. DOFS has been widely adopted in civil, aerospace, and industrial monitoring, where the possibility of collecting information from extended structures is crucial [5], [6], [7]. In this context, optical frequency domain reflectometry (OFDR), based on Rayleigh backscattering, enables the detection and mapping of mechanical strain or temperature variations along the fiber with sub-millimeter spatial resolution [8], [9]. This capability makes OFDR-based DOFS a promising technology for biomedical applications, where high spatial resolution, sensitivity, and reliability are often required [10].

Although FOSs are already well established in several biomedical fields, including minimally invasive surgery, biosensing, biomechanical analysis, and physiological monitoring, the use of DOFS in biomedical systems is still limited [11], [12]. Recent studies have explored DOFS for thermal mapping during ablation procedures, strain monitoring in biological tissues or flexible medical devices, and shape reconstruction of surgical instruments [13], [14], [15], [16]. However, the potential of DOFS for developing extended sensing interfaces able to capture physiological signals from the interaction with the human body remains largely underexploited. This aspect is particularly relevant for contact-based monitoring systems. Physiological and biomechanical activities, such as breathing, heartbeat, posture changes, and body movements, generate forces, pressures, and micro-deformations at the interface between the body and external supports [17]. Monitoring these interactions through an extended sensing surface could provide spatially resolved information that cannot be obtained from single-point sensors.

However, current solutions for distributed or multi-point physiological monitoring mainly rely on electrical sensor arrays, such as capacitive or piezoresistive sensors, or on quasi-distributed FOS configurations based on arrays of fiber Bragg grating sensors (FBGs) [18], [19], [20], [21], [22]. Electrical sensor arrays typically require complex readout electronics and multi-channel wiring, which may increase bulkiness and reduce system flexibility. Conversely, FBG-based systems require predefined sensing points along the fiber, limiting spatial resolution and adaptability to complex geometries. OFDR-based DOFS can overcome several of these limitations by providing high-resolution distributed measurements using a standard optical fiber and a single interrogation unit, without requiring dense electronic networks or predefined grating arrays.

Nevertheless, the direct use of bare optical fibers in body-interfacing systems is limited by their mechanical fragility. For this reason, suitable integration strategies are needed to protect the fiber and ensure efficient transfer of external mechanical perturbations to the sensing element. Previous works have embedded FOSs into metal housings, 3D-printed structures, or polymeric matrices to improve robustness and reliability [23], [24], [25], [26]. In this framework, soft polymeric matrices are particularly suitable for biomedical sensing interfaces because they can protect the fiber while preserving conformability to non-planar surfaces and compatibility with direct body interaction [27], [28].

Based on these considerations, this work presents a soft DOFS-based mat (DM) designed as a distributed sensing interface for cardiorespiratory monitoring. The system consists of a single-mode optical fiber embedded in a soft silicone matrix and arranged in a serpentine layout to cover an extended sensing area. Compared with previous studies in which optical fibers were integrated into silicone matrices for DOFS-based pressure or force mapping, the proposed DM is intended as a large-area body-interfacing platform for physiological monitoring [29], [30]. In this study, the design and fabrication of the proposed DM are presented, together with an exploratory assessment in which the system is placed on the backrest of a standard office chair and used in a seated nearable configuration. The analysis first investigates the mechanical interaction between the subject and the sensing surface through the reconstruction of time-resolved two-dimensional response maps. Then, the distributed signals are processed to extract respiratory and cardiac components and to estimate respiratory rate (RR) and heart rate (HR). Overall, this work aims to demonstrate the potential of the proposed DOFS-based mat as a large-area nearable sensing platform for monitoring physiological activity through body-device mechanical interaction.

## II. DISTRIBUTED OPTICAL FIBER SENSING MAT: WORKING PRINCIPLE, DESIGN AND FABRICATION

### *A. Working principle of DOFS based on OFDR*

DOFS relies on optical scattering phenomena occurring during light propagation inside the fiber, mainly including Raman, Brillouin, and Rayleigh scattering [31]. In this work, the sensing mechanism is based on Rayleigh backscattering, which originates from microscopic refractive index fluctuations along the fiber core. These local inhomogeneities generate a characteristic backscattered profile, commonly referred to as the Rayleigh fingerprint or local spectral signature [32]. When the fiber is subjected to external perturbations, such as mechanical strain or temperature variations, the local optical properties of the fiber change, producing a shift of the Rayleigh signature with respect to a reference state. The analysis of this spectral shift enables distributed localization of the perturbation along the fiber [33].

In an OFDR system, the light backscattered from the sensing fiber is interferometrically combined with the signal from an optical reference path, enabling reconstruction of the Rayleigh backscatter signature along the fiber. The reconstructed response is then divided into local sensing windows, where the signature acquired under measurement conditions is compared by cross-correlation with the corresponding baseline signature to estimate the local spectral shift. The discretization of the fiber into distributed sensing sites defines the spatial sampling step, or gage pitch, which can be approximated as:

$$\Delta z = \frac{L}{N} \tag{1}$$

where $L$ is the interrogated fiber length and $N$ is the number of distributed sensing sites. The Rayleigh spectral shift is sensitive to both axial strain and temperature variations and can be expressed, in linear form, as:

$$\Delta f(z) = S_{\varepsilon}\Delta\varepsilon(z) + S_{T}\Delta T(z) \tag{2}$$

where $\Delta f(z)$ is the local Rayleigh spectral shift, $\Delta\varepsilon(z)$ is the strain variation, $\Delta T(z)$ is the temperature variation, and $S_{\varepsilon}$ and $S_{T}$ are the strain and temperature sensitivity coefficients, respectively [34].

In this study, no dedicated characterization was performed to convert the optical output into absolute strain or temperature values. Therefore, the DOFS output was directly analyzed as local Rayleigh spectral shift, expressed in GHz. Since the tests were conducted under controlled environmental conditions and focused on dynamic cardiorespiratory-induced perturbations, the measured spectral shift was mainly interpreted as an indicator of the distributed mechanical response generated by the interaction between the subject and the sensing mat.

### *B. Design, fabrication and interrogation setup of the DOFS-based mat*

The DM was designed as a soft and large-area sensing interface integrating a single-mode optical fiber within a silicone matrix. The mat has a square geometry of 40 cm × 40 cm, a total thickness of 1 cm, and the optical fiber is embedded at mid-thickness, i.e., 0.5 cm from the upper surface, to promote the transfer of surface mechanical perturbations to the sensing fiber while limiting the overall bulk of the system. DragonSkin™20 silicone rubber (Smooth-On, Inc., USA) was selected as material because of its biocompatibility, flexibility, conformability to non-planar surfaces, and ability to provide mechanical protection to the embedded fiber. The optical fiber was arranged in a continuous serpentine layout to maximize the sensing coverage over the mat surface. The path consists of 18 horizontal segments spaced 2 cm apart, resulting in an embedded fiber length of approximately 7.5 m (see Fig. 1(A)). T-shaped anchoring elements with elastic straps were also integrated into the mat to facilitate stable attachment to non-planar external supports.

The mat was fabricated through a two-step silicone casting process described in Fig. 1(B). First, a modular PLA mold was designed in Autodesk Inventor® (Autodesk Inc., USA) and fabricated by fused deposition modeling using an Ultimaker S5 3D printer (Ultimaker B.V., Netherlands). The mold was designed with raised features reproducing the serpentine fiber path and the cavities for the T-shaped anchoring elements, in order to obtain the corresponding grooves in the first silicone layer. After the first casting and curing, the 0.5 cm-thick silicone layer was removed from the mold and flipped, exposing the grooves generated by the mold features. Then, PLA 3D-printed lateral walls with a height of 1 cm were positioned along the layer perimeter to define the volume for the second casting. Before pouring the second silicone layer, the optical fiber was manually placed inside the serpentine groove, reinforced at the external ends, and locally fixed with small drops of silicone to preserve its layout. The T-shaped anchoring pins and elastic straps were also inserted into the dedicated grooves. Finally, the second silicone layer was cast to fully encapsulate the fiber and anchoring elements. After curing, the lateral walls were removed, obtaining the final 1 cm-thick device.

After fabrication, the DM was connected to the ODiSI 6102 system (Luna Inc., USA), based on OFDR and Rayleigh backscattering, for the measurement setup [35]. The system was configured with a gage pitch of 2.6 mm and a sampling frequency of 50 Hz. After excluding the fiber inlet/outlet portions and the curved regions of the serpentine layout, 2250 sensing sites were obtained over the active fiber length of approximately 5.8 m (see Fig. 1(C)).

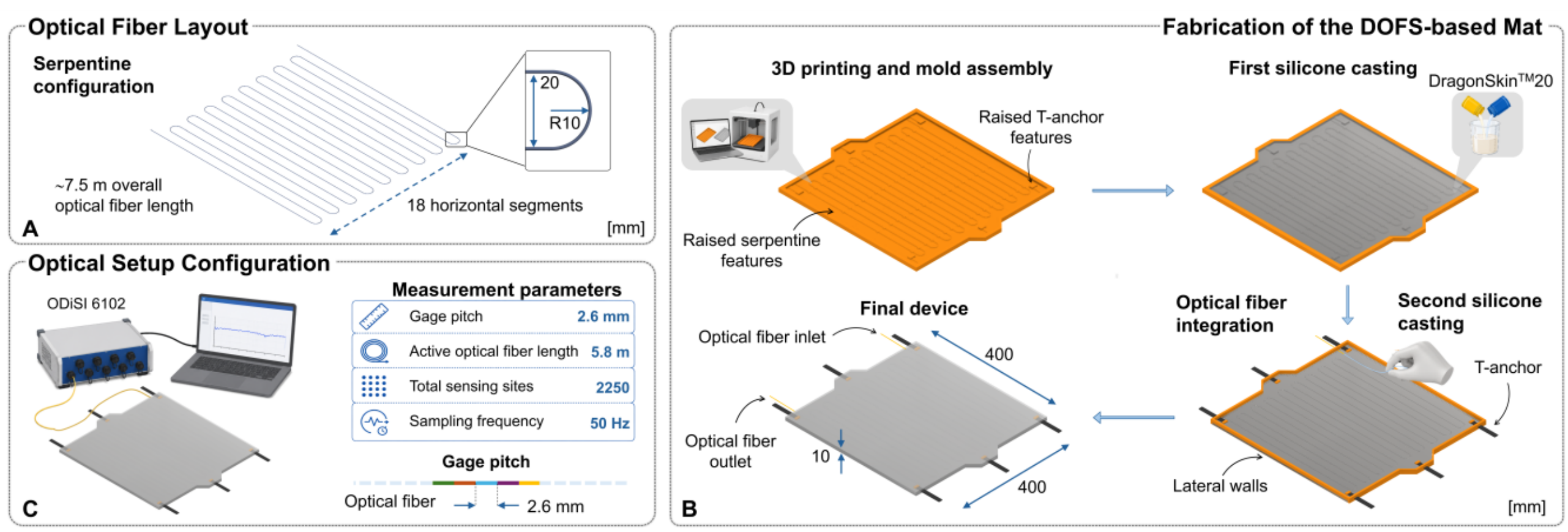

**Fig. 1.** Design, fabrication, and optical interrogation setup of the DOFS-based mat. (A) Serpentine optical fiber layout and main geometrical parameters; (B) fabrication workflow based on 3D-printed molds, two-step silicone casting, and integration of the optical fiber and T-shaped anchors; (C) interrogation setup and main acquisition parameters.

## III. NEARABLE CARDIORESPIRATORY MONITORING IN A SEATED POSITION: SETUP, PROTOCOL AND DATA ANALYSIS

Cardiorespiratory monitoring provides relevant information on physiological status and can support the detection of altered conditions related to stress, fatigue, or respiratory and cardiac dysfunctions [36]. This is particularly relevant in static environments, such as office-like scenarios, where prolonged seated posture and cognitive load may affect physiological responses [37]. In this context, the DM was evaluated in a nearable configuration by placing it on the back of a chair, with the aim of detecting mechanical disturbances induced by the cardiorespiratory system without requiring the subject to wear the detection system. The rationale of the test relies on the mechanical coupling between the subject's back and the sensing surface. During breathing, thoracic expansion and contraction generate periodic body displacements that can be transmitted to the mat and detected by the embedded fiber. Similarly, cardiac activity produces smaller heartbeat-related micro-motions, often associated with ballistocardiographic components, which can modulate the optical response of the sensing fiber [38].

### *A. Experimental setup and protocol*

Six healthy volunteers, three males and three females aged between 24 and 62 years, were recruited for the study. All experimental procedures complied with the Declaration of Helsinki and were approved by the Ethics Committee of Università Campus Bio-Medico di Roma (protocol code: 72.25 CET2cbm ATLETIC, approval date: 27 May 2025). All participants provided informed consent before enrollment. The DM was positioned on the backrest of a standard office chair and secured using the integrated elastic straps and T-shaped anchoring elements. The straps were tensioned and fixed to the back of the chair using high-adhesion kinesiology tape to ensure stable coupling between the mat and the chair during the acquisition. Participants were instructed to sit in an upright posture with their back in contact with the sensing surface. During the entire protocol, they also wore a Zephyr™ BioHarness (BH) chest strap, used as reference system for the simultaneous acquisition of the breathing waveform and electrocardiogram (ECG). Fig. 2(A) shows the experimental setup employed during the trials. Each subject performed a guided breathing protocol composed of the following sequence: 20 s of apnea (A1), 120 s of normal breathing (NB), 20 s of apnea (A2), 30 s of fast breathing (FB), 10 s of apnea (A3), 30 s of slow breathing (SB), and 10 s of final apnea (A4). Signals acquired from the DM, sampled at 50 Hz, and from the BH, sampled at 25 Hz for the breathing waveform and 250 Hz for the ECG, were imported into MATLAB® (MathWorks Inc., USA) for subsequent processing and analysis. Fig. 2(B) shows, by way of example, the signals extracted from selected sensing sites of the DM throughout the experimental protocol for one subject.

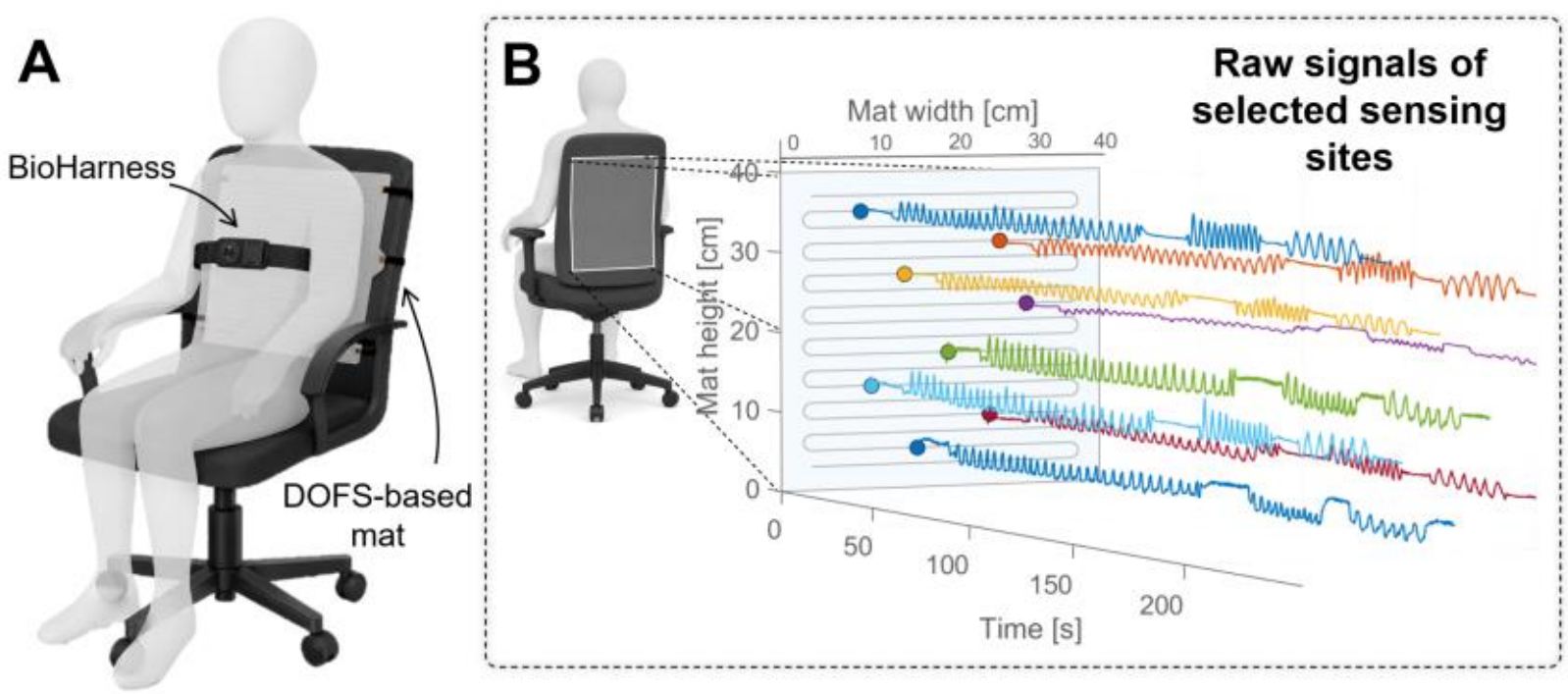

**Fig. 2.** (A) Experimental setup used in the cardiorespiratory monitoring tests; (B) raw signals extracted from selected sensing sites on the mat during the experimental protocol.

### *B. Data analysis*

The raw output acquired from the DM, corresponding to the local spectral shift provided by the OFDR system, was first reorganized according to the serpentine layout of the embedded fiber. Specifically, data were structured into a 3D matrix composed of 18 rows, corresponding to the horizontal fiber segments, 125 columns, corresponding to the sensing sites along each segment, and a third dimension representing time. DM data were temporally synchronized with the signals recorded by the BH, using the beginning of A1 and the end of A4 as temporal markers, and then normalized in the [0, 1] range. A graphical detail on the organization of the distributed DM data into a 3D matrix is provided in Appendix A of the Supplementary Materials. Starting from this common data structure, three analyses were performed to investigate the spatial response of the mat and to estimate RR and HR.

#### *1) Two-dimensional response map reconstruction during cardiorespiratory activity*

Time-resolved two-dimensional maps were generated from the synchronized and normalized DM data to visualize the spatial distribution of the mat response during the experimental protocol. Each map corresponds to one temporal slice of the 3D matrix and provides an instantaneous representation of the spectral shift distribution over the sensing surface. The synchronization with the BH signals allowed the extraction of maps corresponding to specific respiratory phases and representative instants of the cardiac cycle. These maps were used to investigate how cardiorespiratory activity modulates the spatial mechanical interaction between the subject's back and the DM.

#### *2) Respiratory signal extraction and respiratory rate estimation*

For RR estimation, the signals from all 2250 sensing points of the DM (see Fig. 3(A)) and the breathing waveform recorded by the BH were filtered using a first-order Butterworth band-pass filter between 0.05 Hz and 1.5 Hz. The filtered DM signals were then averaged over the entire sensing area to obtain a single respiratory signal representative of the global mat response (see Fig. 3(B)). Both the average DM signal and the BH breathing waveform were segmented into the three respiratory conditions of the protocol: NB, FB, and SB. For each condition, respiratory peaks were detected in the time domain using the *findpeaks* function in MATLAB®. Given two consecutive respiratory peaks occurring at $t_i$ and $t_{i+1}$, the breath-by-breath RR was computed as:

$$RR = \frac{60}{t_{i+1} - t_i} \qquad [apm] \tag{3}$$

where apm indicates respiratory acts per minute. Fig. 3(C) shows the RR values estimated from the DM and BH data for one enrolled subject during the experimental protocol. Additional details on the signal-processing procedures for RR estimation are provided in Appendix C of the Supplementary Materials.

The agreement between DM and BH estimates was evaluated using mean absolute error (MAE), mean absolute percentage error (MAPE), and Bland-Altman analysis. MAE and MAPE were computed as:

$$MAE = \frac{1}{N}\sum_{k=1}^{N}\left|RR_k^{DM} - RR_k^{BH}\right| \qquad [apm] \tag{4}$$

$$MAPE = \frac{1}{N}\sum_{k=1}^{N}\frac{\left|RR_k^{DM} - RR_k^{BH}\right|}{RR_k^{BH}} \cdot 100 \qquad [\%] \tag{5}$$

where $N$ is the number of respiratory intervals within a given condition, and $RR_k^{DM}$ and $RR_k^{BH}$ are the k-*th* RR values estimated from the DM and BH signals, respectively. Regarding Bland-Altman analysis, all RR estimates from the different subjects and respiratory conditions were aggregated into two separate vectors, one for the DM and one for the BH. The corresponding values were then plotted on the $(RR^{DM} - RR^{BH})\ vs. ((RR^{DM} + RR^{BH})/2)$ plane, where the mean of the differences (MOD) and the limits of agreement (LOAs) were also reported.

*3) Cardiac signal extraction and heart rate estimation*

For HR estimation, the analysis was focused on a selected region of the DM located near the thoracic area corresponding to the heart. This region included the sensing sites between the 4*th* and 14*th* horizontal fiber segments and between the 30*th* and 110*th* sensing sites along each segment. The analysis was performed on A1, NB, and A2, which were selected and then segmented because of their lower artifact content. Fig. 4(A) shows the raw signals of the DM with the ECG of the BH extracted for one enrolled subject during NB condition. For the DM, the 891 signals of the selected sensing regions were processed using a third-order Butterworth high-pass filter with a cutoff frequency of 4 Hz, envelope extraction with a 100-sample window, and a first-order Butterworth band-pass filter between 0.7 Hz and 2 Hz. The filtered signals from the selected region were then averaged to obtain a representative cardiac-related signal. The ECG signal recorded by the BH was processed using a third-order Butterworth band-pass filter between 4 Hz and 50 Hz, envelope extraction with a 500-sample window, and the same 0.7 Hz – 2 Hz band-pass filter used for the DM signal. Fig. 4(B) shows both the DM and BH filtered signals for one subject during the NB condition.

HR was estimated in the frequency domain using the Welch method to compute the power spectral density (PSD). Signals were analyzed using overlapping windows with a 1 s sliding step. An 8 s window was used for A1 and A2, while a 30 s window was used for NB. For each window, the frequency corresponding to the maximum PSD peak was identified and converted into HR as:

$$HR = 60 \cdot f_{peak} \qquad [bpm] \tag{6}$$

where $f_{peak}$ is the frequency of the dominant PSD peak and bpm indicates beats per minute. Fig. 4(C) shows the HR values estimated from the DM and BH data for a subject during the NB condition. Additional details on the signal-processing procedures for HR estimation are provided in Appendix C of the Supplementary Materials.

The agreement between HR values estimated from the DM and BH was assessed using MAE, MAPE, and Bland-Altman analysis, as described for RR estimation.

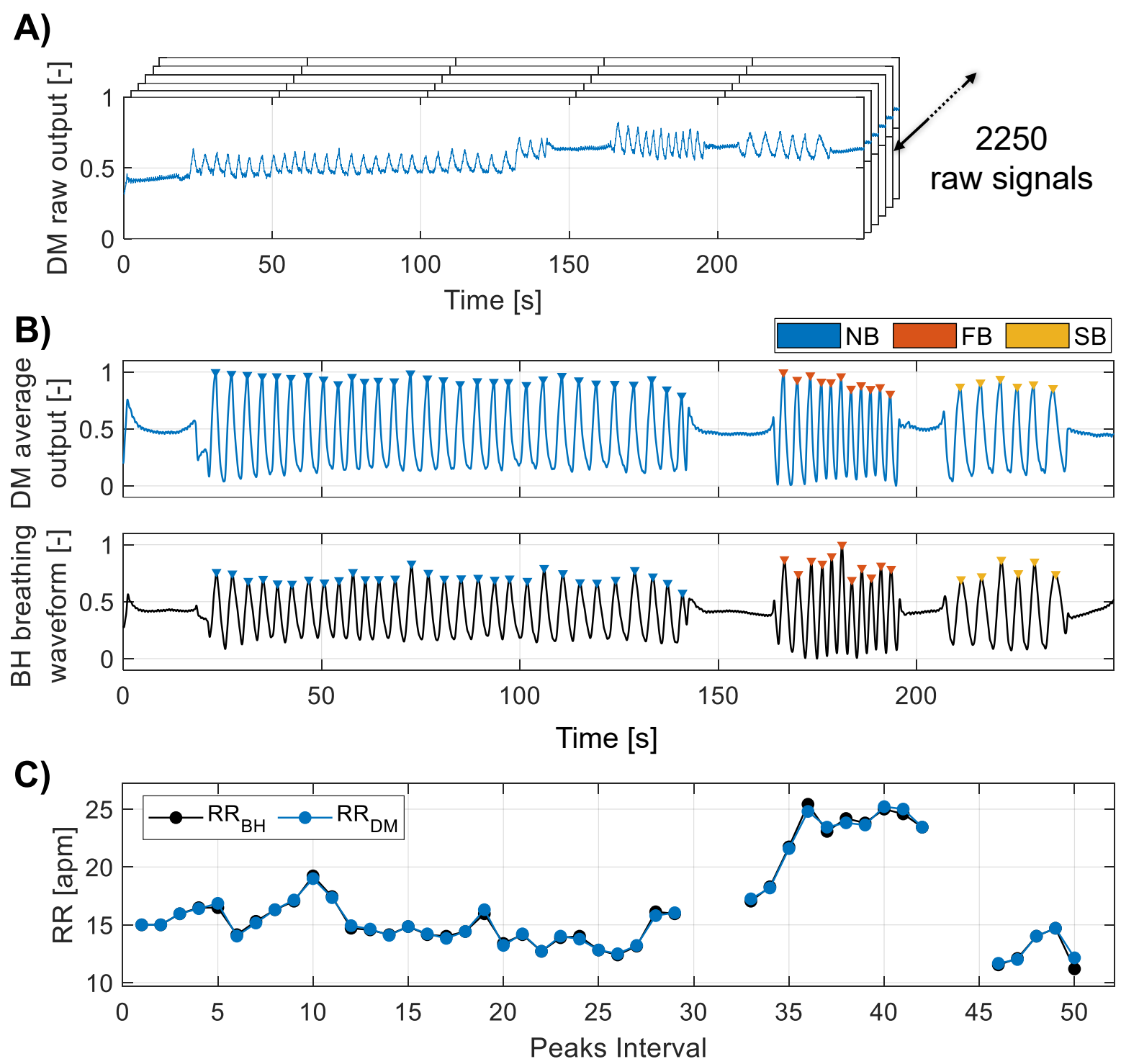

**Fig. 3.** Workflow for respiratory rate extraction: from the raw signals acquired by the DOFS-based mat (A), through peak detection on the averaged mat signal and the reference breathing waveform (B), to the comparison of the respiratory rate estimates provided by the mat and the reference system (C).

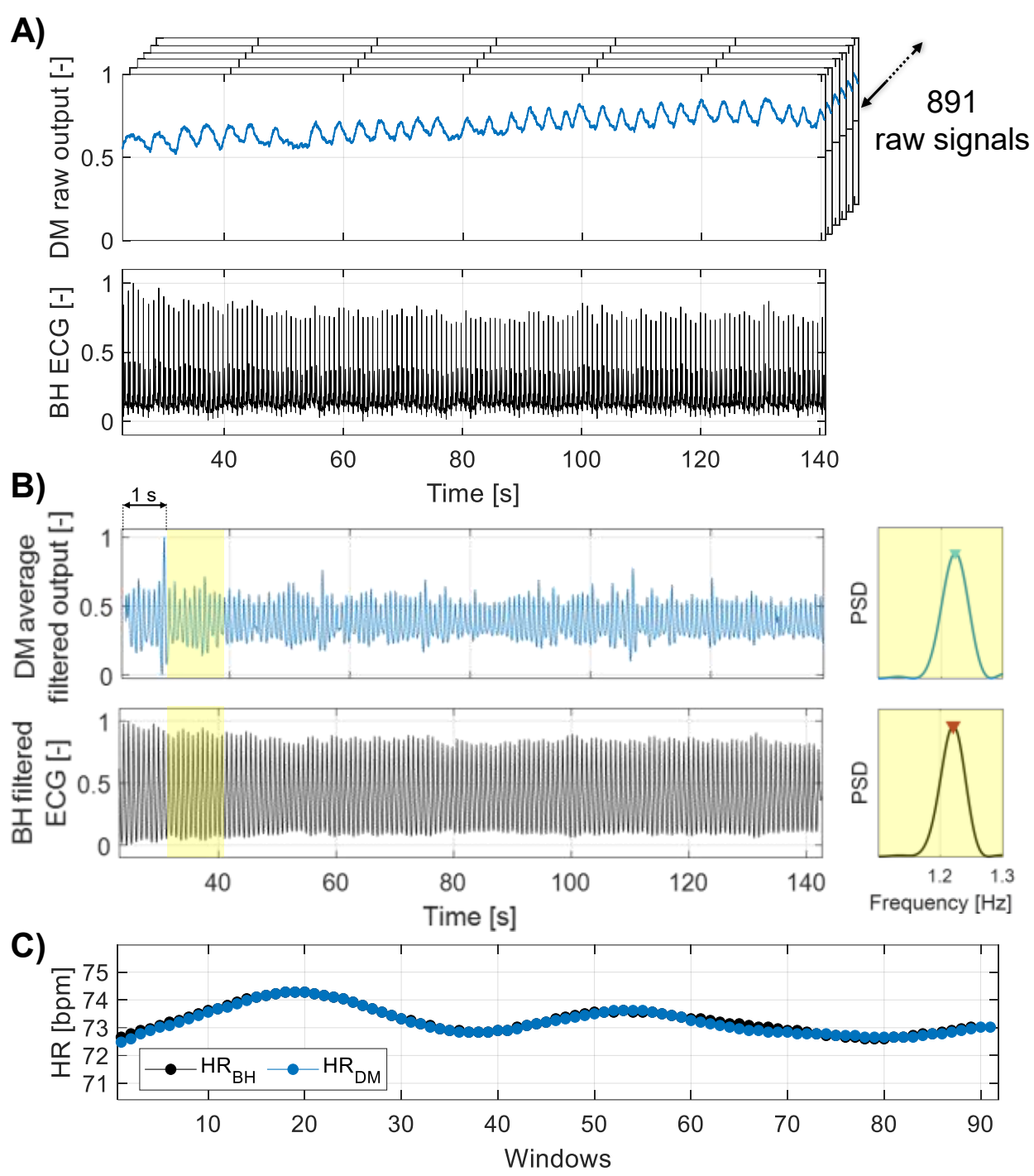

**Fig. 4.** Workflow for heart rate extraction: from the raw signals acquired by the DOFS-based mat (A), through frequency domain analysis on the averaged mat signal and the reference ECG (B), to the comparison of the heart rate estimates provided by the mat and the reference system (C).

## IV. NEARABLE CARDIORESPIRATORY MONITORING IN A SEATED POSITION: RESULTS AND DISCUSSION

### *A. Two-dimensional response map reconstruction during cardiorespiratory activity*

Fig. 5 shows representative two-dimensional maps obtained from one enrolled subject during a single respiratory act identified from the breathing waveform recorded by the BH under NB conditions. The maps reveal a time-varying redistribution of the DM output over the sensing area, reflecting changes in the mechanical interaction between the subject's back and the chair-mounted mat throughout the respiratory cycle. The spatial response is not uniformly distributed across the mat. Higher output levels are mainly observed in the upper-central and central-to-lower regions, which are consistent with the areas where the shoulders, thoracic region, and lower back establish stronger mechanical coupling with the backrest. Conversely, the lateral regions generally exhibit a weaker response. This behavior is reasonably attributable to the curvature of the torso and flanks, which reduces direct contact with the relatively planar sensing surface in a seated posture. The resulting pattern therefore reflects both the subject-specific back-mat contact distribution and the superimposed mechanical perturbations induced by respiration. During inspiration, an increase in the signal can be observed in several of the most strongly coupled regions, consistent with the transmission of thoracic expansion to the sensing surface. During expiration, the response progressively decreases as the respiratory-induced mechanical load is released. Moreover, the location and intensity of the active regions vary over the selected instants, indicating that the respiratory motion is transmitted non-uniformly across the back rather than as a spatially homogeneous displacement. This spatially resolved representation may provide useful information for analyzing individual respiratory mechanics and identifying possible asymmetries or abnormal ventilatory patterns [39].

Representative two-dimensional maps associated with cardiac activity, together with the corresponding qualitative discussion, are reported in Appendix B of the Supplementary Materials.

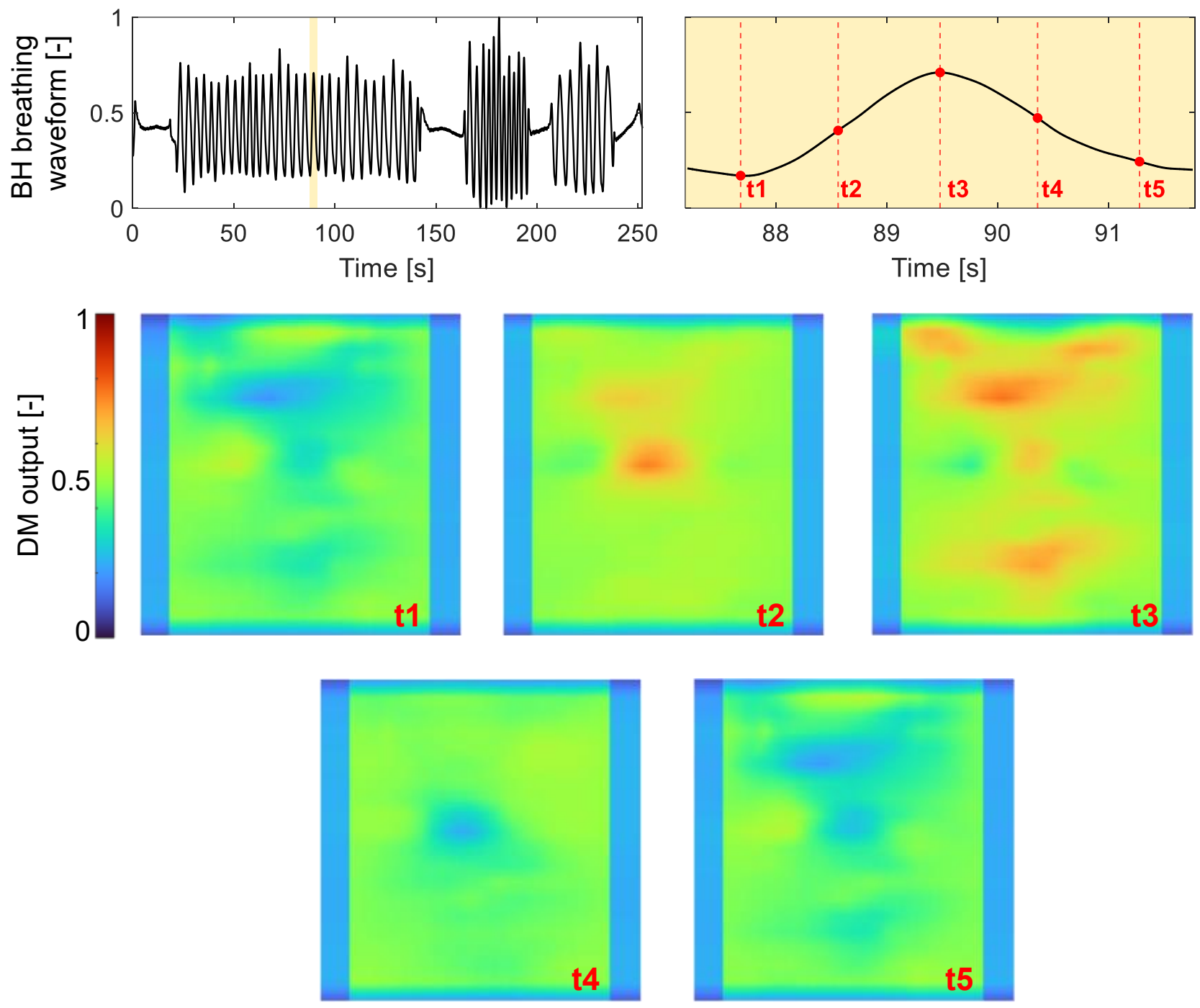

**Fig. 5.** BioHarness breathing waveform with selected time instants t1–t5 with corresponding two-dimensional maps of the normalized mat output.

### *B. Respiratory signal extraction and respiratory rate estimation*

The RR estimates obtained from the DM were compared with the reference values provided by the BH. The resulting MAE and MAPE values, together with the Bland-Altman assessment of agreement between the two systems, are shown in Fig. 6(A).

Overall, the DM provided reliable RR estimates across the different respiratory conditions. For most subjects and conditions, the MAE remained below 1 apm. The main exception was observed for subject S5, for whom the MAE reached 1.24 apm during SB and exceeded 5 apm during FB. The same subject also showed the highest MAPE values, while increased errors were also observed for S2 and S6 during SB. Apart from these cases, MAPE values remained generally below 4%, confirming a good correspondence between the DM and BH estimates. The Bland-Altman analysis further supported this agreement, with a MOD equal to 0.04 apm and LOAs ranging from −3.91 apm to 3.98 apm. These values indicate a limited bias and acceptable variability between the RR estimates obtained from the DM and the BH.

These results are consistent with previous works on non-invasive respiratory monitoring, although direct comparison is affected by differences in sensing technology, experimental protocol, and measurement configuration [40], [41], [42]. For example, FBG-based mattresses and under-mattress systems have demonstrated high accuracy for respiratory monitoring in lying configurations [40], [43]. However, such configurations generally benefit from a more stable mechanical coupling between the body and the sensing surface. In contrast, the present study considered a vertical, non-planar, and less constrained configuration, where the transmission of respiratory-induced movements to the sensing mat is more susceptible to posture, contact variability, and motion artifacts.

Other studies have proposed nearable systems mounted on the backrest of chairs for respiratory monitoring [44], [45], [46]. These systems share a similar application scenario but are generally based on smaller sensing elements or quasi-distributed configurations. The DM proposed in this work covers a larger sensing area and exploits the distributed nature of OFDR-based sensing, allowing respiratory information to be extracted from the global response of the mat rather than from predefined sensing points only. This feature may improve robustness against subject positioning and partial back–mat contact, which are typical issues in seated nearable monitoring.

Additional details on the results for RR estimation are provided in Appendix C of the Supplementary Materials.

### *C. Cardiac signal extraction and heart rate estimation*

The HR estimates obtained from the DM were compared with the reference values provided by the BH. The resulting MAE and MAPE values, together with the Bland-Altman assessment of agreement between the two systems, are shown in Fig. 6(B). The comparison with the BH showed that the DM was able to estimate HR with good accuracy in the analyzed conditions, namely A1, NB, and A2. Across all subjects and conditions, MAE values remained below 3.21 bpm, while MAPE values were below 5%. Bland-Altman analysis showed a mean of the differences equal to −0.56 bpm, with limits of agreement ranging from −6.12 bpm to 5.01 bpm. These values indicate a limited bias and acceptable variability between the HR estimates obtained from the DM and the reference system.

The obtained HR performance is in line with previous studies on non-invasive cardiac monitoring using fiber-optic and other unobtrusive nearable sensing technologies, although these systems were generally evaluated with subjects in a lying position [47], [48], [49]. In contrast, detecting cardiac activity in a seated nearable configuration is particularly challenging [46]. Heartbeat-induced mechanical displacements are much smaller than respiratory movements and can be easily masked by posture-related changes, residual breathing components, and variability in the body–sensor contact. Moreover, compared with lying configurations, the seated posture provides a less stable and less uniform mechanical coupling between the thoracic region and the sensing surface, thereby reducing the transmission of cardiac-induced micro-motions.

Despite these challenges, the results indicate that the DM can detect cardiac-related components and provide HR estimates consistent with the BH reference. This finding is relevant because, unlike single-point sensors, the proposed distributed approach does not require the cardiac signal to be captured at a predefined location. Instead, the extended sensing surface can be exploited to identify regions where heartbeat-induced mechanical components are more evident. This represents a promising feature for future optimization of nearable DOFS-based cardiorespiratory monitoring systems.

Additional details on the results for HR estimation are provided in Appendix C of the Supplementary Materials.

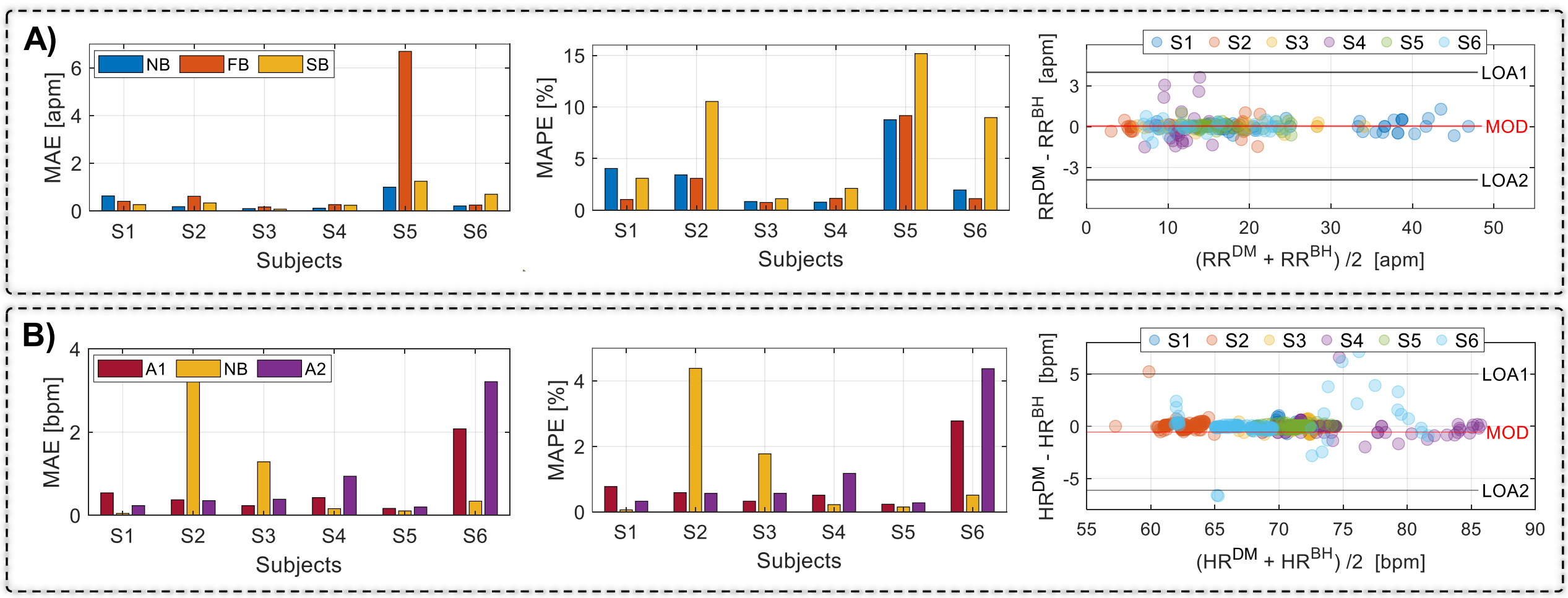

**Fig. 6.** Graphical performance assessment of the DOFS-based mat: subject- and condition-specific MAE and MAPE values, together with Bland-Altman analysis, for (A) respiratory rate estimation and (B) heart rate estimation.

## V. CONCLUSIONS

This work presented the design, fabrication, and experimental assessment of a DOFS-based mat conceived as a soft, large-area, nearable sensing interface for cardiorespiratory monitoring. The system integrates a single-mode optical fiber within a silicone matrix according to a serpentine layout, enabling the distributed acquisition of local spectral shifts over the mat surface through OFDR interrogation. Soft encapsulation, complete fiber integration, and embedded anchoring elements resulted in a flexible platform suitable for stable attachment to non-planar supports, such as the backrest of a standard office chair.

The mat was assessed in a seated nearable configuration, in which it was not worn by the subject but operated as a body-interfacing surface for detecting the mechanical perturbations generated by cardiorespiratory activity. The distributed measurements enabled the reconstruction of time-resolved two-dimensional maps of the back-mat interaction, providing spatial information on respiratory-induced thoracic motion and weaker cardiac-related components. This capability represents a key strength of the proposed approach, as it moves beyond single-point and quasi-distributed sensing by providing a spatially resolved representation of physiological mechanical activity. The pilot assessment also showed that the DM could estimate both RR and HR in agreement with the BH reference system, despite the challenging seated configuration, where body-system coupling is less stable and uniform than in lying configurations.

The present signal-processing approach should nevertheless be considered an initial strategy that does not yet exploit the full spatial information provided by the 2250 sensing sites. RR and HR were estimated from signals obtained by averaging the response of the entire mat or relatively large sensing regions. Although this approach ensured methodological consistency across subjects, the resulting average signals may also include sensing sites characterized by weak physiological components, poor mechanical coupling, noise, or motion-related artifacts. The distributed nature of the DM offers the possibility of overcoming this limitation by identifying the sites or spatial regions providing the highest-quality respiratory and cardiac signals.

Future work will focus on spatially adaptive processing strategies capable of identifying the sensing sites with the highest physiological information content while excluding poorly informative or noise-dominated responses. By combining only the most representative sites, these methods could provide cleaner respiratory and cardiac signals and improve the accuracy, robustness, and inter-subject consistency of RR and HR estimation. Broader validation on larger populations and under different postural conditions will also be required to assess the generalizability of the proposed approach.

## ACKNOWLEDGMENT

This article reflects only the authors' views and opinions, and neither European Union nor European Commission can be considered responsible for them.

# Supplementary Materials

## Appendix A. Data organization into a 3D matrix

The first step of the analysis involved reorganizing the DM raw data (i.e., local spectral shift obtained from the OFDR system). Specifically, the signals were structured into a 3D matrix composed of 18 rows (corresponding to the horizontal segments of the serpentine-shaped optical fiber) and 125 columns (representing the number of sensing sites per horizontal segment). The third dimension of the matrix thus represents the temporal evolution of the measurements. Thus, for each time instant, signals from 18 x 125 = 2250 sensing sites were obtained. Figure 1 provides a schematic representation of the 3D matrix structure, illustrating the spatial organization of the sensing sites across the DM and their correspondence with the temporal signals.

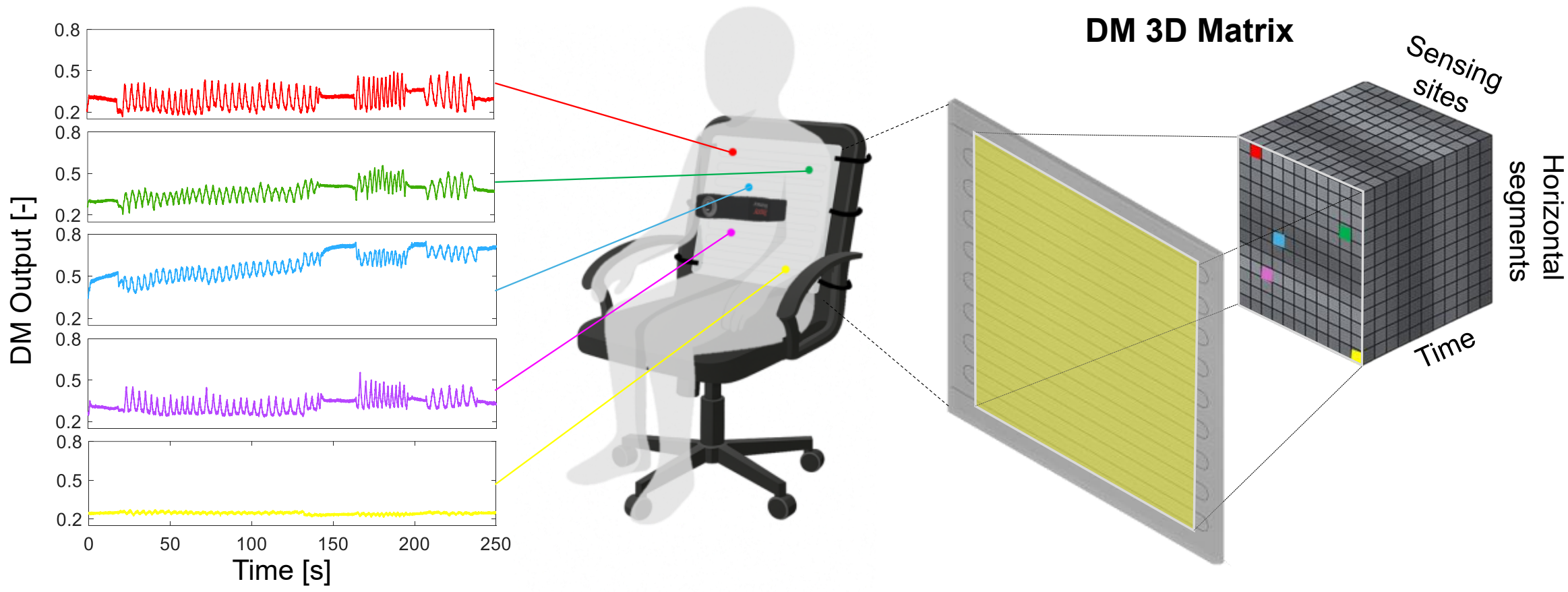

**Figure 1.** Conceptual representation of the 3D data matrix generated from the DM, with rows corresponding to horizontal fiber segments, columns to sensing sites, and depth to time samples.

## Appendix B. Two-dimensional response map reconstruction during cardiorespiratory activity

Based on synchronized and normalized data from the BH and DM, time-resolved heatmaps of the DM output were generated to visualize the spatially distributed mechanical response associated with both the contact between the volunteer's back and the chair backrest and the superimposed cardiorespiratory activity. Each heatmap corresponds to a temporal slice of the previously described 3D matrix and therefore provides an instantaneous representation of the DM response at a specific time point. Synchronization with the signals acquired by the BH also enabled the identification of instants corresponding to specific respiratory phases and characteristic moments of the cardiac cycle, allowing the extraction of the associated two-dimensional maps. These maps were used to qualitatively investigate how physiological events are reflected in the spatial response patterns captured by the DM.

Only a few previous studies have reported two-dimensional maps obtained using optical fiber-based systems interrogated by OFDR and integrated into silicone substrates for specific application scenarios. In particular, [1] and [2] presented spatial maps associated with pressure- and force-related mechanical interactions under controlled experimental conditions. In the present study, instead, the maps were reconstructed to analyze the distributed response induced by interaction with the human body in a more dynamic and variable scenario. Furthermore, whereas previous studies mainly focused on spatial mapping under controlled mechanical loading, the proposed approach correlates the distributed DM output with physiological reference signals acquired from the BH, namely the ECG and breathing waveform. This enables the qualitative interpretation of how respiratory and cardiac activity modulates the mechanical interaction between the subject and the sensing surface.

Figure 2(A) shows a representative sequence of heatmaps extracted during a single respiratory act identified from the reference breathing waveform acquired by the BH under NB conditions. The two-dimensional distribution of the DM output highlights the regions of the subject's back where the mechanical coupling with the mat is more pronounced. During inspiration, an increase in the DM output is observed across several sensing regions, consistent with the transmission of thoracic expansion to the sensing surface. Conversely, during expiration, the signal progressively decreases as the respiratory-induced mechanical interaction is reduced. This representation, combining spatial and temporal information, may support the qualitative analysis of individual respiratory mechanics and provide a basis for identifying asymmetries or regional differences associated with abnormal ventilatory patterns [3].

Figure 2(B) shows a representative sequence of heatmaps corresponding to different time points within a single cardiac cycle identified from the ECG signal recorded by the BH during condition A1. The maps reveal a broader, relatively stable response pattern associated with the static mechanical coupling between the subject's back and the mat during breath-hold. Within this pattern, smaller localized variations can be observed over time, consistent with heartbeat-induced micro-motions transmitted to the sensing surface. As expected, these cardiac-related variations are less pronounced than the respiratory-induced changes.

Detecting cardiac-related mechanical components is inherently more complex because their amplitude is lower and they can be masked by slower and larger contributions, including residual respiratory motion and posture-related variations. It should also be noted that the maps were generated from the raw DM output without applying dedicated filtering strategies to isolate cardiac components. More specific signal-processing approaches could therefore improve the readability of the maps and facilitate the extraction of cardiac-related spatial information. Moreover, the mechanical response is physiologically delayed with respect to the electrical activity recorded by the ECG because of the electromechanical delay between myocardial depolarization and mechanical contraction [4].

Despite these limitations, the obtained maps provide preliminary evidence that the DM can capture localized and time-varying mechanical components associated with cardiac activity. This distributed representation may support future investigations of cardiac-related spatial patterns and the development of nearable approaches for cardiorespiratory monitoring [5].

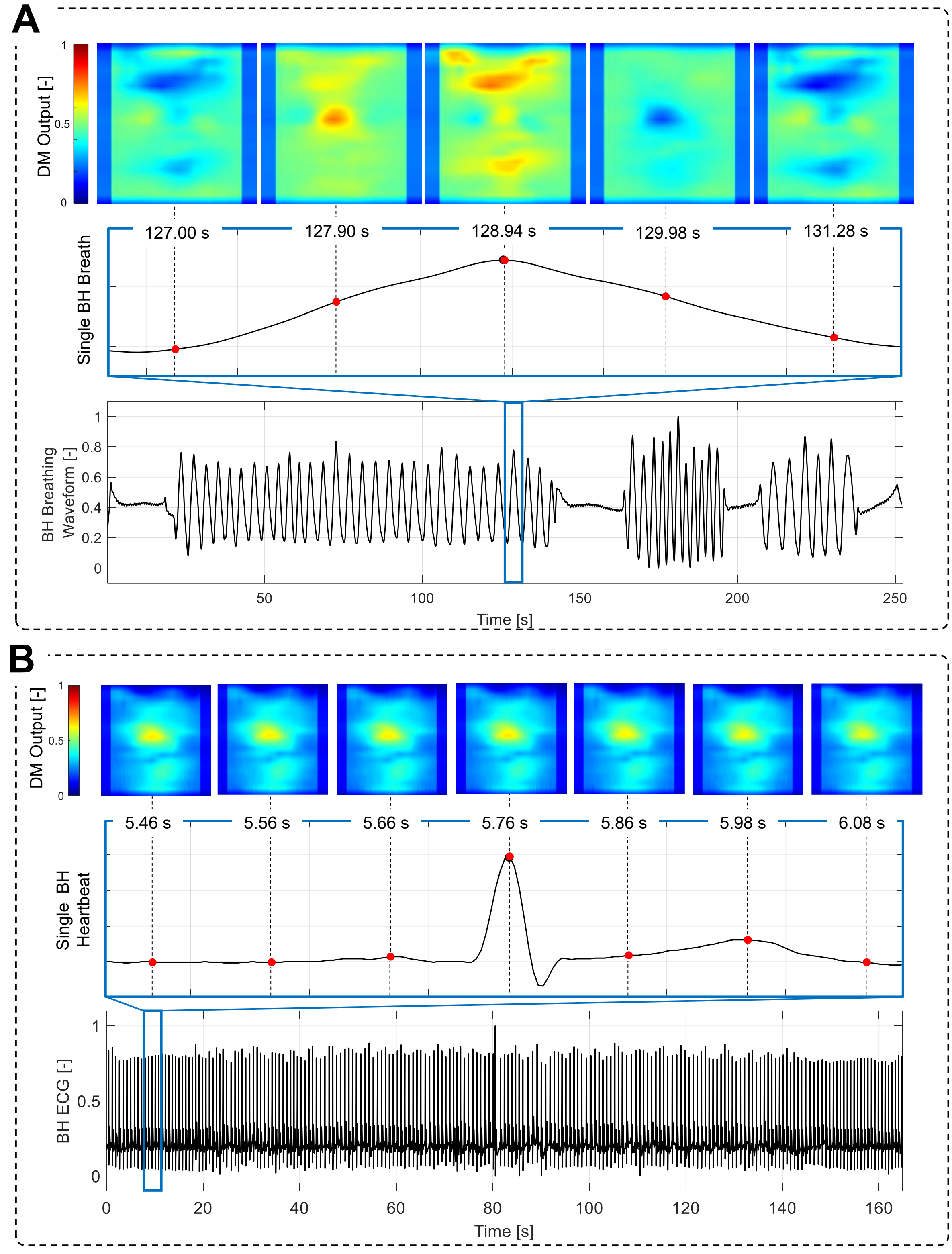

**Figure 2.** (A) Time instants extracted from a representative respiratory act identified on the reference (BH) breathing waveform, with corresponding two-dimensional maps of the normalized output obtained from the DOFS-based mat (DM). (B) Time instants extracted from a representative cardiac cycle identified on the BH ECG signal, with corresponding two-dimensional maps of the normalized DM output.

## Appendix C. More on data analysis and results for respiratory and heart rates estimation

Separate analyses were then performed to independently estimate respiratory rate (RR) and heart rate (HR). To ensure methodological consistency across all enrolled subjects, an estimation strategy based on the spatial averaging of selected sensing sites was adopted, enabling the extraction of robust global signals representative of the dominant respiratory and cardiac dynamics. The following sections describe in detail the specific processing steps implemented and the main results obtained for RR and HR estimation, respectively.

### Respiratory rate estimation: data analysis and results

To estimate RR, the signals from the 2250 sensing sites of the DM and the corresponding breathing waveform from the BH were analyzed. All signals were filtered using a first-order Butterworth band-pass filter with cutoff frequencies of 0.05 Hz and 1.5 Hz. The filtered signals from the mat were then reorganized into a new 3D filtered matrix and averaged to obtain a single signal representative of the entire DM surface. Figure 3(A) shows the main signal processing steps applied to the raw data from the DM for RR estimation.

Both the average DM signal and the breathing waveform from the BH were segmented into three respiratory conditions: NB, FB, and SB. For each condition, a time-domain analysis was performed to estimate RR based on the identification of respiratory peaks using the findpeaks function in MATLAB®. Given the $t_i$ and $t_{i+1}$ associated with two consecutive respiratory peaks, the breath-by-breath RR was computed as:

$$RR = \frac{60}{t_{i+1} - t_i} \qquad [apm] \tag{1}$$

where apm indicates respiratory acts per minute (see Figure 3(B)). To evaluate the accuracy of the DM compared to the BH reference, the mean absolute error (MAE) and mean absolute percentage error (MAPE) were computed as follows:

$$MAE = \frac{1}{N} \sum_{k=1}^{N} \left| RR_k^{DM} - RR_k^{BH} \right| \qquad [apm] \tag{2}$$

$$MAPE = \frac{1}{N} \sum_{k=1}^{N} \frac{\left| RR_k^{DM} - RR_k^{BH} \right|}{RR_k^{BH}} \cdot 100 \qquad [\%] \tag{3}$$

where $N$ is the number of respiratory intervals within a given condition, and $RR_k^{DM}$ and $RR_k^{BH}$ denote the breath-by-breath RR estimates obtained from the DM and the BH for the k-th interval between two consecutive respiratory peaks, respectively. Finally, to assess the agreement between the RR estimates obtained with the DM and those obtained with the BH, a Bland-Altman analysis was performed [6]. For this purpose, all RR estimates from the different subjects and respiratory conditions were aggregated into two separate vectors, one for the DM and one for the BH. The corresponding values were then plotted on the $(RR^{DM} - RR^{BH})\ vs. ((RR^{DM} + RR^{BH})/2)$ plane, where the mean of the differences (MOD) and the limits of agreement (LOAs) were also reported.

The results of RR estimation are reported in terms of MAE and MAPE in Table 1 and graphically illustrated with the Bland-Altman plot in Figure 3(C). When compared to the reference values provided by the BH, all subjects and respiratory conditions showed an MAE below 1 apm, with the exception of subject S5. In this case, the MAE reached 1.24 apm in the SB condition (still within acceptable limits) and exceeded 5 apm during FB. Regarding MAPE, the highest values were also observed for S5, and to a lesser extent for S2 and S6 during SB. For all other subjects, MAPE remained below 4% across all respiratory conditions. Overall, Bland-Altman analysis revealed a good agreement between the two RR estimation methods: the MOD value of 0.04 apm indicates a negligible bias between $RR^{DM}$ and $RR^{BH}$, while the LOA interval (–3.91 apm, 3.98 apm) suggests low variability between the measurements.

Without excluding outliers or subjects with noisy signals, the results still align well with existing literature, despite differences in experimental protocols and sensing technologies [7], [8], [9], [10], [11], [12]. For example, the study [7] proposes a mattress that integrates an FBG array for respiratory monitoring during sleep. Experimental tests demonstrated an MAE and MAPE of less than 0.4 apm and 1.5%, and MOD and LOAs values of 0.01 apm and  0.74 apm, respectively, under normal and fast breathing conditions. Another study proposes a ferroelectric sensor that can be placed under the bed mattress for respiratory monitoring [12]. Experimental tests reported an MAE of 0.59 apm and LOAs of -2.5 apm and 2.2 apm for normal breathing. In these cases, however, the proposed systems are usually used in the supine position and placed on rigid surfaces, which helps to improve the acquisition of the respiratory signal. In contrast, in this study, the DM was installed vertically on a non-rigid, non-planar surface, an inherently more challenging configuration due to increased mechanical deformation and motion artifacts.

Other works have described nearable systems incorporating sensors encapsulated in silicone matrices or 3D-printed components, mounted on the backrest of chairs [13], [14], [15]. For example, the study [13] proposes a sensing element integrating an FBG into a silicone, which during experimental tests demonstrated a MAPE of less than 1% under normal and fast breathing conditions. The study [14], on the other hand, proposes the use of an array of FBGs to be arranged on pillows or sponges to be placed on the back of an office chair, which has demonstrated through experimental tests good performance in respiratory monitoring. These studies propose the positioning of nearables in a manner similar to DM. However, these systems are much smaller in size than DM and therefore, although more functional and less prone to motion artifacts, can only work for specific subject positions.

The DM proposed in this study thus proves effective for non-invasive RR monitoring, with additional benefits stemming from the distributed nature of the sensing approach. This feature enables reliable estimation even in suboptimal postural conditions or when the back–mat contact is limited to a small portion of the sensing surface, overcoming several limitations typical of current state-of-the-art configurations.

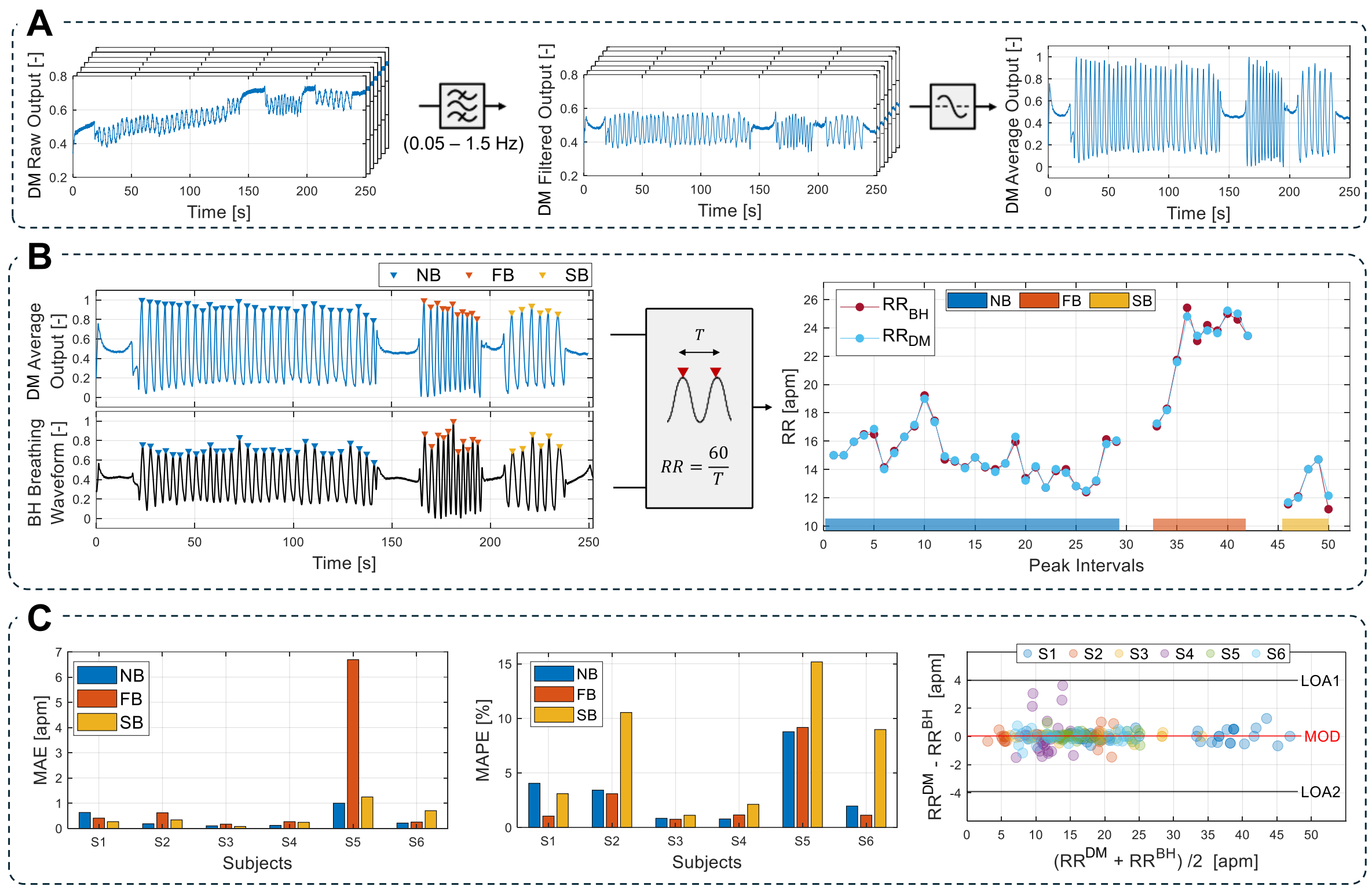

**Figure 3.** (A) Main signal processing steps applied to the raw data from the DOFS-based mat (DM) for respiratory rate (RR) estimation. (B) Detected respiratory peaks on the filtered average signals from the DM and the reference device (BH) during the different phases of the experimental protocol, along with the corresponding RR values. (C) Results of RR estimation in terms of MAE, MAPE, and Bland-Altman plot comparing DM and BH values.

**Table 1. MAE and MAPE for Respiratory Rate Estimation.**

| | MAE [apm] | | | MAPE [%] | | |
|---|---|---|---|---|---|---|
| Subject | NB | FB | SB | NB | FB | SB |
| S1 | 0.63 | 0.41 | 0.27 | 4.04 | 1.04 | 3.10 |
| S2 | 0.18 | 0.62 | 0.34 | 3.42 | 3.08 | 10.53 |
| S3 | 0.10 | 0.17 | 0.07 | 0.83 | 0.74 | 1.11 |
| S4 | 0.12 | 0.26 | 0.24 | 0.77 | 1.14 | 2.11 |
| S5 | 0.99 | 6.69 | 1.24 | 8.76 | 9.16 | 15.17 |
| S6 | 0.21 | 0.25 | 0.70 | 1.95 | 1.12 | 8.97 |

**Heart rate estimation: data analysis and results**

To estimate HR, a portion of the DM located near the thoracic region corresponding to the heart area was selected. Specifically, all sensing sites between the 4th and 14th horizontal fiber segments and between the 30th and 110th measurement sites along each segment were considered. As a reference, the ECG signal provided by the BH was used. The analysis focused on phases A1, NB, and A2 of the experimental protocol, selected due to their low artifact content. For this reason, as shown in Figure 4(A), the 3D matrix of the DM was divided into three 3D submatrices, one for each condition considered.

All signals underwent three pre-processing steps. For the DM, a third-order Butterworth high-pass filter with a cutoff frequency of 4 Hz was applied, followed by upper envelope extraction using a 100-sample window, and finally a first-order Butterworth band-pass filter with cutoff frequencies of 0.7 Hz and 2 Hz. For each of the three conditions (i.e., A1, NB and A2), the filtered signals from the selected mat region were then rearranged into a 3D filtered matrix and averaged to obtain a single signal representative of the entire sensing area. Figure 4(B) shows as an example the processing of the DM signals extracted in the NB condition. For the ECG signal, preprocessing involved a third-order Butterworth band-pass filter between 4 Hz and 50 Hz, followed by envelope detection using a 500-sample window, and finally application of the same band-pass filter (0.7 Hz – 2 Hz) used for the DM signals [16].

Considering the average DM and BH signals for the three experimental conditions, a frequency-domain analysis was performed using the Welch method to estimate the power spectral density (PSD), implemented in MATLAB® [17]. Signals were segmented into overlapping windows with a 1-second sliding step: A1 and A2 were analyzed with an 8 s window, while NB used a 30 s window. For each window, the frequency corresponding to the maximum PSD peak was identified. This frequency was then converted to beats per minute (bpm) by multiplying its value by 60 to obtain HR.

The results related to HR estimation are reported in terms of MAE and MAPE in Table 2 and are graphically represented with the Bland-Altman plot in Figure 4(C). The comparison with the reference values obtained from the BH shows that, for all subjects and across all experimental conditions (i.e., A1, NB, and A2), the MAE remains consistently below 3.21 bpm. Similarly, MAPE values are below 5% for all subjects, confirming the strong accuracy of the HR estimates provided by the DM. In addition, Bland-Altman analysis revealed a MOD value of -0.56 bpm and a range of LOAs between -6.12 bpm and 5.01 bpm, suggesting very good agreement in HR estimation of DM with BH, with negligible bias and relatively low variability between measurements.

Consistent with what was observed for RR estimation, the performance metrics obtained for HR are also in line with those reported in the literature, despite differences in measurement technologies and experimental protocols [18], [19], [20], [21], [22], [23], [24]. These findings confirm the effectiveness of the proposed system, highlighting several advantages over more conventional approaches. Notably, the DM does not require precise positioning of the subject, as is often the case with single point sensors, since the entire surface of the system can potentially contribute to detecting the cardiac signal. Moreover, compared to sensorized mats described in the literature, the DM in this study was evaluated in a seated position, which generally poses a greater challenge for cardiac signal detection. In lying position, body weight helps maintain a stable and uniform contact between the thoracic cage and the sensing system, facilitating the transmission of the micro-displacements induced by cardiac activity. In contrast, in the seated position, the contact is less constrained and less uniform, leading to attenuation of the mechanical signal acquired.

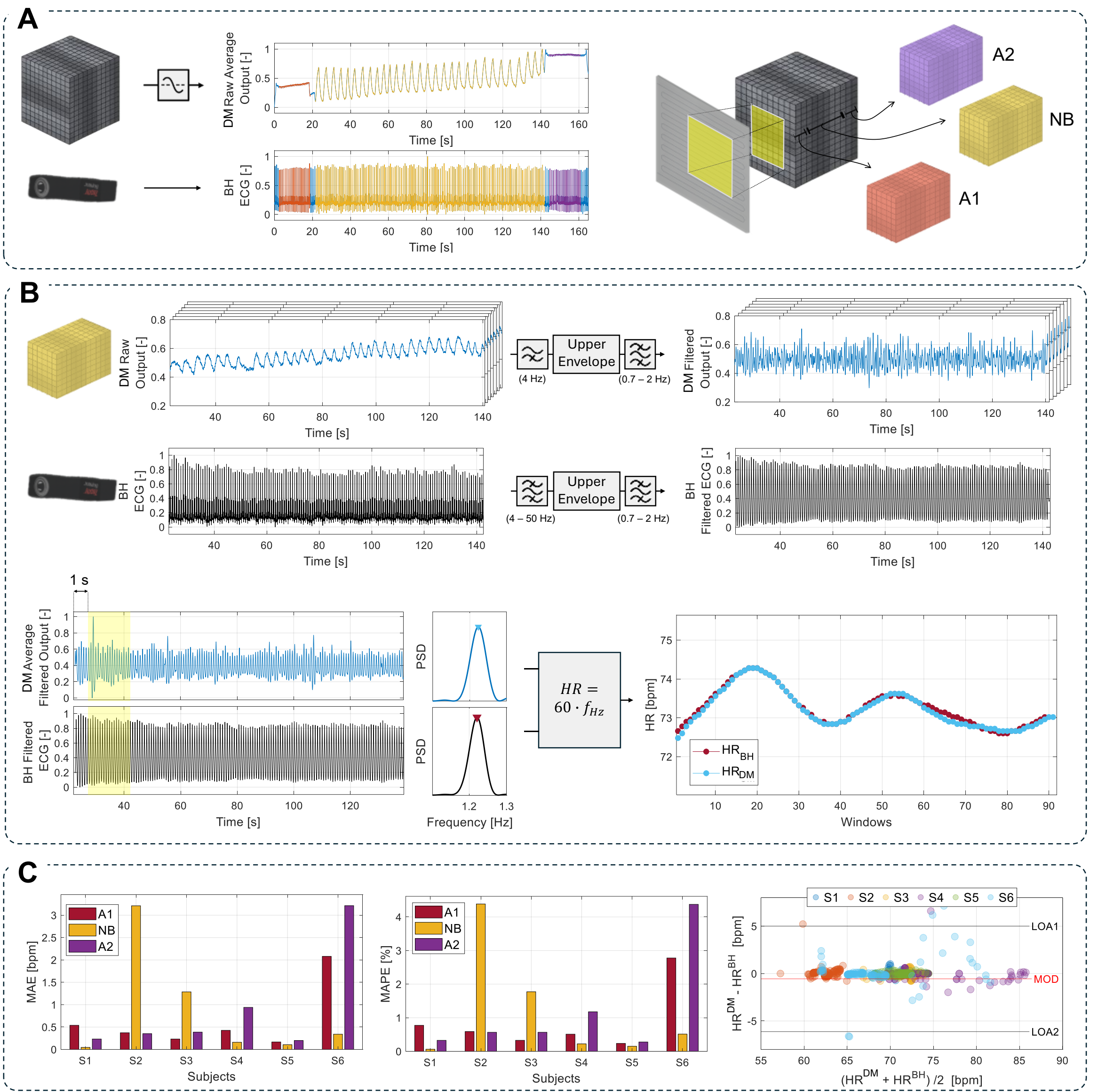

**Figure 4.** (A) Extraction of signal portions from the DOFS-based mat (DM) and reference (BH) data for heart rate (HR) analysis, highlighting the segmentation of the DM 3D matrix into three sub-matrices corresponding to the different phases of the experimental protocol (A1, NB, A2). (B) Main steps of the signal processing pipeline applied to both DM and BH data during the NB phase, including frequency domain analysis and temporal trend of HR estimation. (C) HR estimation results in terms of MAE, MAPE, and Bland-Altman plot comparing DM and BH values.

**Table 2. MAE and MAPE for Heart Rate Estimation.**

| | MAE [bpm] | | | MAPE [%] | | |
|---|---|---|---|---|---|---|
| Subject | A1 | NB | A2 | A1 | NB | A2 |
| S1 | 0.54 | 0.05 | 0.23 | 0.77 | 0.06 | 0.33 |
| S2 | 0.37 | 3.21 | 0.35 | 0.59 | 4.39 | 0.57 |
| S3 | 0.23 | 1.29 | 0.39 | 0.33 | 1.77 | 0.58 |
| S4 | 0.43 | 0.16 | 0.94 | 0.51 | 0.22 | 1.18 |
| S5 | 0.17 | 0.11 | 0.2 | 0.24 | 0.15 | 0.28 |
| S6 | 2.08 | 0.34 | 3.21 | 2.78 | 0.51 | 4.37 |